\documentclass{article}

\usepackage{amsmath}
\usepackage{latexsym}
\usepackage{amssymb}
\usepackage{amsbsy}
\usepackage{epsfig}
\usepackage{url}

\newtheorem{theorem}{Theorem}
\newtheorem{lemma}[theorem]{Lemma}

\newtheorem{ex}[theorem]{Example}
\newtheorem{defn}[theorem]{Definition}
\newtheorem{proposition}[theorem]{Proposition}
\newcommand{\qed}{\hspace{\stretch{1}}$\Box$}
\newenvironment{proof}{\begin{quote}\textbf{Proof:}}{\hfill\qed\end{quote}}
\newenvironment{example}{\begin{ex}\rm}{\hfill\qed\end{ex}}

\long\def\ignore#1{}

\long\def\pjs#1{}
\long\def\jeff#1{}
\long\def\jl#1{}
\long\def\wh#1{}

\long\def\short#1{}
\long\def\noproofs#1{#1}
\long\def\noreified#1{}

\newcommand{\VV}{{\cal V}}
\newcommand{\ZZ}{{\cal Z}}
\newcommand{\RR}{{\cal R}}

\newcommand{\DD}{{\cal D}}
\newcommand{\bnd}{\mathit{bnd}}

\newcommand{\ibounds}{bounds($\ZZ$)}
\newcommand{\rbounds}{bounds($\RR$)}
\newcommand{\dbounds}{bounds($\DD$)}
\newcommand{\Ibounds}{Bounds($\ZZ$)}

\newcommand{\Dbounds}{Bounds($\DD$)}

\newcommand{\range}[2]{\left[\,#1\,..\,#2\,\right]}

\newcommand{\rangeop}{\operatorname{range}}

\newcommand{\bequiv}{\stackrel{b}{\equiv}}

\newcommand{\dom}{\mathit{dom}}

\newcommand{\vars}{\mathit{vars}}

\newcommand{\bigsqcap}{\mathop{\lower.1ex\hbox{\Large$\sqcap$}}}

\newcommand{\calldiff}{\mbox{\upshape\ttfamily alldifferent}}
\newcommand{\lin}{\mathit{lin}}

\def\eclipse{{ECL$^i$PS$^e$}}

\begin{document}

\title{Finite Domain Bounds Consistency Revisited}
\author{
C.W.~Choi\footnote{Department of Computer Science and Engineering,
The Chinese University of Hong Kong, Shatin, N.T., Hong Kong SAR, China.
Email: \texttt{\{cwchoi,jlee\}@cse.cuhk.edu.hk}}
\and 
W.~Harvey\footnote{IC-Parc, William Penney Laboratory, Imperial College London, Exhibition Road, London SW7 2AZ. Email: \texttt{wh@icparc.ic.ac.uk}} 
\and 
J.H.M.~Lee\footnotemark[1]
\and 
P.~J.~Stuckey\footnote{NICTA Victoria Laboratory, Department of Computer Science \& Software Engineering, University of Melbourne, 3010, Australia. Email: \texttt{pjs@cs.mu.oz.au}}
}
\date{\today}
\maketitle


\begin{abstract}
A widely adopted approach to solving constraint satisfaction problems 
combines systematic tree search with constraint propagation
for pruning the search space. 
Constraint propagation is performed by propagators implementing a certain
notion of consistency.
Bounds consistency is the method of choice for
building propagators for arithmetic constraints 
and several global constraints in the finite integer domain.
However, there has been some confusion in the 
definition of bounds consistency.
In this paper we clarify the differences and similarities
among the three commonly used notions of bounds consistency.
\end{abstract}


\section{Introduction}\label{sec:intro}

Finite domain constraint programming combines 
backtracking tree search with constraint propagation
to solve {\em constraint satisfaction problems\/} (CSPs)~\cite{mackworth}.
This framework is realized in constraint programming systems, 
such as \eclipse{}~\cite{eclipse-intro}, SICStus Prolog~\cite{sics}
and ILOG Solver~\cite{solver52}, which 
have been successfully applied to many real-life industrial 
applications.

\emph{Constraint propagation}, based on local consistency algorithms, removes 
infeasible values from the domains of variables to reduce the
search space.
The origins of this technique in the Artificial Intelligence
field concentrated on consistency notions for
constraints with two or fewer variables.
The most successful consistency technique 
was \emph{arc consistency}~\cite{mackworth}
which ensures that for each constraint, every value in the domain
of one variable has a supporting value in the
domain of the other variable which satisfies the constraint.

Arc consistency can be naturally extended to
constraints of more than two variables. This
extension has been called \emph{generalized arc consistency}~\cite{gac4},
as well as \emph{domain consistency}~\cite{pdv2}
(which is the terminology we will use),
and \emph{hyper-arc consistency}~\cite{book}.
Checking domain consistency is NP-complete
even for linear equations, an important kind of constraint.

To avoid this problem, weaker forms of consistency were introduced
for handling constraints with large numbers of variables.
The most notable one for linear arithmetic constraints
has been \emph{bounds consistency} 
(sometimes called \emph{interval consistency}).
Unfortunately there are \emph{three} commonly used but \emph{incompatible}
definitions of bounds consistency in the literature.
This is confusing to practitioners in the 
design and implementation of efficient bounds consistency algorithms,
as well as for users of constraint programming system claiming to support
bounds consistency.
In this paper we clarify the three existing definitions of bounds consistency 
and the differences between them. 
We also gives conditions under which the definitions may coincide 
for certain constraints.

\section{Background}\label{sec:backgrounds}

In this paper we consider integer constraint solving
using constraint propagation.
Let $\ZZ$ denote the integers, and $\RR$ denote the reals.
Hence $\ZZ \models F$ denotes that the formula $F$
is universally true in the integers 
(that is when the constants,
functions and  variables
are interpreted in the integer domain).
Similarly $\RR \models F$ denotes that the formula $F$
is universally true in the reals.

We consider a given (finite) set of integer variables $\VV$,
which we shall sometimes interpret as real variables.
Each variable is associated with a finite set of possible values,
defined by the domain.
A \emph{domain} $D$ is a complete mapping 
from a set of variables $\VV$ to finite sets of integers.
We assume in this paper that $D(v)$ for all $v \in \VV$ are totally ordered.

A \emph{valuation} $\theta$ is a mapping of 
variables to values (integers or reals), written 
$\{x_1 \mapsto d_1, \ldots, x_n \mapsto d_n\}$.
Let $\vars$ be the function that returns the set of
variables appearing in an expression, constraint or valuation.
Given an expression $e$, $\theta(e)$ is obtained by replacing 
each $v \in \vars(e)$ by $\theta(v)$ and calculating
the value of the resulting variable free expression.

In an abuse of notation, we define a valuation 
$\theta$ to be an element of a domain $D$, 
written $\theta \in D$, iff $\theta(v) \in \ZZ$ 
and $\theta(v) \in D(v)$ for all $v \in \vars(\theta)$.
We are interested in determining the infimums and supremums of
expressions with respect to some domain $D$.
Define the \emph{infimum} and \emph{supremum} of an 
expression $e$ with respect to a
domain $D$ as $\inf_D e = \inf\{ \theta(e) ~|~ \theta \in D\}$ 
and $\sup_D e = \sup\{ \theta(e) ~|~ \theta \in D\}$.
We also use \emph{range} notation: $\range{l}{u}$
denotes the set $\{ d \in \ZZ ~|~ l \leq d \leq u \}$ 
when $l$ and $u$ are integers.
A domain is a \emph{range domain} if $D(x)$ is a range for all $x$.
Let $D' = \rangeop(D)$ be the smallest range domain containing $D$, 
i.e.\ domain $D'(x) = \range{\inf_D x}{\sup_D x}$ for all $x \in \VV$.

A constraint places restriction on the allowable values for
a set of variables and is usually written in well understood 
mathematical syntax. More formally, a {\em constraint\/}  
$c$ is a relation expressed using available function and relation 
symbols in a specific constraint language.  
For the purpose of this paper, we assume the usual integer 
interpretation of arithmetic constraints and logical operators 
such as $\neg$, $\wedge$, $\vee$, $\Rightarrow$, and $\Leftrightarrow$.
We call $\theta$ an \emph{integer solutions} of $c$ iff
$\vars(\theta) = \vars(c)$ and $\ZZ \models_\theta c$.
Similarly, we call $\theta$ a \emph{real solutions} of $c$ iff
$\vars(\theta) = \vars(c)$ and $\RR \models_\theta c$.
A \emph{constraint satisfaction problem} (CSP) consists 
of a set of constraints read as conjunction.

Constraint propagation enforcing domain consistency ensures that 
for each constraint $c$, every value in the domain of each variable
can be extended to an assignment satisfying $c$.

\begin{defn}\label{defn:dom}
A domain $D$ is \emph{domain consistent} for a constraint $c$
where $\vars(c) = \{x_1, \ldots, x_n\}$,
if for each variable $x_i, 1 \leq i \leq n$ and 
for each $d_i \in D(x_i)$ there exist
integers $d_j$ with $d_j \in D(x_j)$,
$ 1 \leq  j \leq n, j \neq i$ such that
$\theta = \{x_1 \mapsto d_1, \ldots, x_n \mapsto d_n\}$ is an 
integer solution of $c$.
\end{defn}

\begin{example} \label{ex:1}
Consider the constraint $c_{\lin}~~ \equiv ~~ x_1 = 3 x_2 + 5 x_3$.
The domain $D_0$ defined by 
$D_0(x_1) = \range{2}{7}$,  
$D_0(x_2) = \range{0}{2}$, and 
$D_0(x_3) = \range{-1}{2}$ is not domain consistent w.r.t.\ $c_{\lin}$.
But the domain $D_1$ defined by 
$D_1(x_1) = \{3,5,6\}$,  
$D_1(x_2) = \{0,1,2\}$, and 
$D_1(x_3) = \{0,1\}$ is domain consistent for $c_{\lin}$,
supported by the integer solutions
$\theta_1 = \{x_1 \mapsto 3, x_2 \mapsto 1, x_3 \mapsto 0\}$,
$\theta_2 = \{x_1 \mapsto 5, x_2 \mapsto 0, x_3 \mapsto 1\}$, and
$\theta_3 = \{x_1 \mapsto 6, x_2 \mapsto 2, x_3 \mapsto 0\}$.
\end{example}

\section{Different Notions of Bounds Consistency}\label{sec:bc}

The basis of bounds consistency is to relax the consistency requirement
to apply only to the lower and upper bounds of the domain of
each variable. 
There are three incompatible definitions of bounds consistency
used in the literature, all for constraints with finite integer domains.
The first and second definitions are based on \emph{integer} solutions,
while the third definition is based on \emph{real} solutions.

\begin{defn}\label{defn:dbounds}
A domain $D$ is \emph{\dbounds{} consistent} for a constraint $c$
where $\vars(c) = \{x_1, \ldots, x_n\}$,
if for each variable $x_i, 1 \leq i \leq n$ and 
for each $d_i  \in \{\inf_D x_i, \sup_D x_i\}$ there exist
\textbf{integers} $\boldsymbol{d_j}$ with $\boldsymbol{d_j \in D(x_j)}$,
$ 1 \leq  j \leq n, j \neq i$ such that
$\theta = \{x_1 \mapsto d_1, \ldots, x_n \mapsto d_n\}$ is an 
\textbf{integer solution} of $c$.
\end{defn}

\begin{defn}\label{defn:ibounds}
A domain $D$ is \emph{\ibounds{} consistent} for a constraint $c$
where $\vars(c) = \{x_1, \ldots, x_n\}$,
if for each variable $x_i, 1 \leq i \leq n$ and 
for each $d_i  \in \{\inf_D x_i, \sup_D x_i\}$ there exist
\textbf{integers} $\boldsymbol{d_j}$ with 
$\boldsymbol{\inf_D x_j \leq d_j \leq \sup_D x_j}$,
$1 \leq  j \leq n, j \neq i$ such that
$\theta = \{x_1 \mapsto d_1, \ldots, x_n \mapsto d_n\}$ is an 
\textbf{integer solution} of $c$.
\end{defn}

\begin{defn}\label{defn:rbounds}
A domain $D$ is \emph{\rbounds{} consistent} for a constraint $c$
where $\vars(c) = \{x_1, \ldots, x_n\}$,
if for each variable $x_i, 1 \leq i \leq n$ and 
for each $d_i  \in \{\inf_D x_i, \sup_D x_i\}$ there exist
\textbf{real numbers} $\boldsymbol{d_j}$ with 
$\boldsymbol{\inf_D x_j \leq d_j \leq \sup_D x_j}$, 
$1 \leq  j \leq n, j \neq i$ such that
$\theta = \{x_1 \mapsto d_1, \ldots, x_n \mapsto d_n\}$ 
is a \textbf{real solution} of $c$.
\end{defn}

Definition~\ref{defn:dbounds} is used in for example for 
the two definitions in
Dechter~\cite[pages 73 \& 435]{rina-book};
Frisch \bgroup \em et al.\egroup~\cite{toby-cp02};
and implicitly in 
Lallouet \bgroup \em et al.\egroup~\cite{arnaud}.
Definition~\ref{defn:ibounds} is far more widely used appearing in 
for example 
{Van Hentenryck} \bgroup \em et al.\egroup~\cite{pdv2};
Puget~\cite{puget-alldifferent};
R{\'e}gin \& Rueher~\cite{sum-diff};
Quimper \bgroup \em et al.\egroup~\cite{vanBeek-gcc}; and
SICStus Prolog~\cite{sics}.
Definition~\ref{defn:rbounds} appears in for 
example 
Marriott \& Stuckey~\cite{book};
Schulte \& Stuckey~\cite{SchulteStuckey-PPDP2001};
Harvey \& Schimpf~\cite{warwick-joachim}; and 
Zhang \& Yap~\cite{ZhangYap00}.
Apt~\cite{apt-book} gives both 
Definitions~\ref{defn:ibounds}~and~\ref{defn:rbounds}, 
calling the former interval consistency 
and the latter bounds consistency.

Let us now examine the differences of the definitions.
\jeff{defined stronger properly}
Consider two notions of consistency $\Omega$ and $\Pi$.
We say that $\Omega$ consistency is \emph{at least as strong as} 
$\Pi$ consistency iff, given any constraint $c$ and domain $D$, 
if $D$ is $\Omega$ consistent for $c$ then $D$ is $\Pi$ consistent
for $c$. We say that $\Omega$ consistency is \emph{stronger than}
$\Pi$ consistency iff $\Omega$ consistency is at least 
as strong as $\Pi$ consistency, but $\Pi$ consistency
is \emph{not} at least as strong as $\Omega$ consistency.
The following relationship between the three notions of bounds
consistency is clear from the definition.

\wh{For some domains the relationship is not strict, so don't claim it is}
\jeff{removed the words stringent condition}
\begin{proposition}\label{prop:bounds_impl}
\Dbounds{} consistency is \emph{stronger} 
than \ibounds{} consistency,
which is stronger than \rbounds{} consistency.
\end{proposition}
\noproofs{
\ignore{
\begin{proof}
(Sketch) Essentially if a domain $D$ is \dbounds{} consistent
for a constraint $c$, then $D$ is also \ibounds{} consistent 
for $c$. Similarly if $D$ is \ibounds{} consistent for $c$,
$D$ is also \rbounds{} consistent for $c$.
\end{proof}
}
\begin{proof}
For each of the notions of consistency, for each bound of each
variable a support must be found, i.e.\ an assignment of values
to each of the other variables.
For each of these other variables, 
let $S_{\DD}$, $S_{\ZZ}$, and $S_{\RR}$ be the set of allowed values to choose 
from for \dbounds{}, \ibounds{} and \rbounds{} consistency respectively.

From the definition, $S_{\DD} \subseteq S_{\ZZ}$,
thus having a support 
for \dbounds{} consistency implies having a
support for \ibounds{} consistency.
The reverse is not always true: see Example~\ref{ex:bounds}.
Hence, \dbounds{} consistency is stronger than \ibounds{} consistency.
Similarly, $S_{\ZZ} \subseteq S_{\RR}$,
so \ibounds{} consistency is stronger than
\rbounds{} consistency (see Example~\ref{ex:bounds} again for a case
that is \rbounds{} consistent but not \ibounds{} consistent).
\end{proof}
}

\begin{example}\label{ex:bounds}
Consider the constraint $c_{\lin}$ from Example~\ref{ex:1}.
The domain $D_2$ defined by 
$D_2(x_1) = \{2,3,4,6,7\}$,  
$D_2(x_2) = \range{0}{2}$, and 
$D_2(x_3) = \range{0}{1}$
is \rbounds{} consistent
(but not \dbounds{} consistent nor \ibounds{} consistent)
w.r.t.\ $c_{\lin}$, 
supported by the real solutions 
$\{x_1 \mapsto 2, x_2 \mapsto 2/3, x_3 \mapsto 0\}$,
$\{x_1 \mapsto 7, x_2 \mapsto 2, x_3 \mapsto 1/5\}$,
$\{x_1 \mapsto 5, x_2 \mapsto 0, x_3 \mapsto 1\}$.

The domain $D_3$ defined by  
$D_3(x_1) = \{3,4,6\}$,  
$D_3(x_2) = \range{0}{2}$, and 
$D_3(x_3) =\range{0}{1}$ 
is \ibounds{} and \rbounds{} consistent 
(but not \dbounds{} consistent)
w.r.t.\ $c_{\lin}$.

The domain $D_4$ defined by 
$D_4(x_1) = \{3,4,6\}$,
$D_4(x_2) = \range{1}{2}$, and
$D_4(x_3) = \{0\}$
is \dbounds{}, \ibounds{} and \rbounds{} consistent 
w.r.t.\ $c_{\lin}$.
\end{example}

The relationship between the \ibounds{} and \dbounds{} consistency
is straightforward to explain.

\begin{proposition}\label{prop:ibdb}
$D$ is \ibounds{} consistent with $c$
iff $\rangeop(D)$ is \dbounds{} consistent with $c$.
\end{proposition}

The second definition of bounds consistency 
in Dechter~\cite[page 435]{rina-book} 
works only with 
range domains.  By Proposition~\ref{prop:ibdb}, the definition
coincides with both \ibounds{} and \dbounds{} consistency.  
Similarly, Apt's~\cite{apt-book} interval consistency is also 
equivalent to \dbounds{} consistency.
But in general, finite domain constraint solvers 
do not always operate on range domains, 
but rather use a mix of propagators implementing
different kinds of consistencies, both domain and bounds consistency.

\begin{example}
Consider the constraint $c_{\lin}$ and domain $D_3$ 
from Example~\ref{ex:bounds}. 
Now $\rangeop(D_3)$ is both \dbounds{} and \ibounds{} consistent 
with $c_{\lin}$.
But, as noted in Example~\ref{ex:bounds},
$D_3$ is only \ibounds{} consistent but \emph{not} \dbounds{} 
consistent with $c_{\lin}$.
\end{example}

Both \rbounds{} and \ibounds{} consistency depend only
on the upper and lower bounds of the domains of the variables 
under consideration.

\begin{proposition}\label{prop:range}
For $\alpha = \RR$ or $\alpha = \ZZ$ and constraint $c$,
$D$ is bounds($\alpha$) consistent for $c$ 
iff $\rangeop(D)$ is bounds($\alpha$) consistent for $c$.
\end{proposition}

This is not the case for \dbounds{} consistency, which
suggests that, strictly, it is not really a form of \emph{bounds consistency}.
Indeed, most existing implementations of bounds propagators 
make use of Proposition~\ref{prop:range}
to avoid reexecuting a bounds propagator unless the lower
or upper bound of a variable involved in the propagator changes.

\begin{example}\label{ex:gap}
Consider the constraint $c_{\lin}$ and domain $D_3$ 
from Example~\ref{ex:bounds} again.
Both $D_3$ and $\rangeop(D_3)$ are \ibounds{} and \rbounds{}
consistency with $c_{\lin}$, but only $\rangeop(D_3)$ is
\dbounds{} consistent with $c_{\lin}$.
\end{example}

There are significant problems with the
stronger \ibounds{} (and \dbounds{}) consistency. 
In particular, for linear equations it is NP-complete to check
\ibounds{} (and \dbounds{}) consistency,
while for \rbounds{} consistency it is only linear time
(e.g.\ see Schulte \& Stuckey~\cite{SchulteStuckey-PPDP2001}).

\begin{proposition}\label{prop:nphard}
Checking \ibounds{}, \dbounds{}, or domain consistency
of a domain $D$ with a linear equality
$a_1 x_1 + \cdots a_n x_n = a_0$
where $\{a_0, \ldots, a_n\}$ are integer constants
and $\{x_1, \ldots, x_n\}$ are integer variables, is NP-complete.
\end{proposition}
\begin{proof}
It is clear that this problem belongs to NP.
Next, we show that the SUBSET SUM NP-complete problem~\cite{clr}
is polynomial reducible to this problem.

\ignore{
The SUBSET SUM problem is: given a set of $n$ integers
$A = \{a_1, \ldots, a_n\}$ and a constant $a_{n+1}$, determine if there
is a set $S \subseteq A$ where 
$\sum \{ a_i ~|~ a_i \in S \wedge 1 \leq i \leq n\} = a_{n+1}$.
For each integers $a_i$ where $1 \leq i \leq n+1$ we associate an
integer variable $x_i$ with the domain $D$ defined by
$D(x_j) = \{0,1\}$ for all $1 \leq j \leq n$ and $D(x_{n+1}) = \{1\}$. 
Now, checking whether the linear equation
$c \equiv a_1 x_1 + \cdots + a_n x_n - a_{n+1} x_{n+1} = 0$
is domain consistent with domain $D$ for variable $x_{n+1}$ 
is to find a valuation $\theta \in D$
where $\theta(x_{n+1}) = 1$ such that $c$ is satisfied.
The valuation $\theta$ defines a set 
$S = \{ a_i ~|~ \theta(x_i) = 1 \wedge 1 \leq i \leq n \}$ where 
$\sum \{ a_i ~|~ a_i \in S \wedge 1 \leq i \leq n \} - a_{n+1} = 0$.
}

In the subset sum problem, we are given a set of $n$ natural numbers
$A = \{a_1, \ldots, a_n\}$ and a target natural number $a_{n+1}$, and we
must determine whether or not there exists a subset $S \subseteq A$ where 
$\sum \{ a_i ~|~ a_i \in S \wedge 1 \leq i \leq n\} = a_{n+1}$.
For each integer $a_i$ where $1 \leq i \leq n+1$ we associate an
integer variable $x_i$ with the domain $D$ defined by $D(x_i) = \{0,1\}$.
We also introduce another integer variable $x_{n+2}$ where $D(x_{n+2}) = \{0,1\}$.
Now, the linear equation $c$ defined by
\[
a_1 x_1 + \cdots + a_n x_n - a_{n+1} x_{n+1} - (\sum_{i=1}^{n} a_i) x_{n+2} = 0
\]
has integer solutions:
\[
\begin{array}{c}
\{x_1 \mapsto 0, \ldots, x_n \mapsto 0, x_{n+1} \mapsto 0, x_{n+2} \mapsto 0\} \\
\{x_1 \mapsto 1, \ldots, x_n \mapsto 1, x_{n+1} \mapsto 0, x_{n+2} \mapsto 1\}
\end{array}
\]
\wh{Spelled out a few more steps in the proof}%
These two solutions exhibit that all domain elements for all variables have
support, with the exception of $x_{n+1} \mapsto 1$.
Thus, $D$ is domain consistent for $c$ iff there exists
a valuation $\theta \in D$ where $\theta(x_{n+1}) = 1$ that satisfies $c$.
Clearly we cannot have $\theta(x_{n+1}) = 1$ and $\theta(x_{n+2}) = 1$ (no
solution), so any such $\theta$ must have $\theta(x_{n+2}) = 0$.
Such a valuation, if it exists, defines a set 
$S = \{ a_i ~|~ \theta(x_i) = 1 \wedge 1 \leq i \leq n \}$ such that
$\sum \{ a_i ~|~ a_i \in S \wedge 1 \leq i \leq n \} = a_{n+1}$;
conversely, the existence of such a set implies a corresponding valuation.
Thus there exists a solution to the SUBSET SUM problem iff there exists a
solution of $c$ with $\theta(x_{n+1}) = 1$ and $\theta(x_{n+2}) = 0$, iff
$D$ is domain consistent with respect to $c$.

For domain $D$, checking \ibounds{}, \dbounds{}, or domain consistency
for $c$ are equivalent. Hence, the cases for checking  \ibounds{} 
and \dbounds{} consistency follow analogously.
\end{proof}

\ignore{
\begin{example}
Consider an instance of the SUBSET SUM problem: 
given a set of integers $A = \{1, 2, 8\}$, 
determine if there is a set $S \subseteq A$ where the sum 
of the integers in $S$ is $3$.

Following the proof, we construct a 
linear equation $c \equiv 1(x_1) + 2(x_2) + 8(x_3) - 3(x_4) = 0$
with domain $D$ defined by $D(x_1) = D(x_2) = D(x_3) = \{0,1\}$
and $D(x_4) = \{1\}$.
We know $c$ is domain consistent with $D$ for $x_4$ since there exists
$\theta = \{ x_1 \mapsto 1, x_2 \mapsto 1, x_3 \mapsto 0, x_4 \mapsto 1\}$
which satisfies $c$ and $\theta(x_4) = 1$.
By collecting those $a_i$'s where $\theta(x_i) = 1$ for $1 \leq i \leq 3$,
we obtain a subset $\{1,2\}$ with the sum $3$.
\end{example}
}

There are constraints, however, for which \rbounds{} consistency
is less meaningful than other forms of consistency.

\begin{example}
Consider the global constraint 
$\calldiff(x_1,x_2,x_3) \equiv
x_1 \neq x_2 \wedge x_1 \neq x_3 \wedge x_2 \neq x_3$
and the domain $D_5$
where 
$D_5(x_1) = D_5(x_2) = \range{1}{2}$ and
$D_5(x_3) = \range{2}{3}$.
Then this domain is not \ibounds{} consistent, since
there is no integer solution with $x_3 = 2$. 
But the domain $D_5$ is \rbounds{} consistent
with $\calldiff(x_1,x_2,x_3)$
since it has real solutions
$\{ x_1 \mapsto 1, x_2 \mapsto 2, x_3 \mapsto 3\}$,
$\{ x_1 \mapsto 2, x_2 \mapsto 1, x_3 \mapsto 3\}$,
$\{ x_1 \mapsto 1, x_2 \mapsto 1.5, x_3 \mapsto 2\}$.
\end{example}

A problem with \rbounds{} consistency
is that it may not
be clear how to interpret an integer constraint in the reals.

\ignore{
\begin{example}
Consider the constraint $x_1 = x_2!$ ($x_1$ is the factorial of $x_2$)
and the domain
$D_6(x_1) = D_6(x_2) = \range{0}{18}$,
then \ibounds{} consistency will demand the reduction
of domains to 
$D_7(x_1) = \range{1}{6}$, $D_7(x_2) = \range{0}{3}$.
Unfortunately to apply
\rbounds{} consistency 
we need to clarify what it means 
to be a real solution of $x_1 = x_2!$.
\end{example}
}
\begin{example}
Consider the constraint $x_1 = x_2 \mod x_3$.
\wh{Cut unimportant details of example}
\ignore{
The domain $D_6$ defined by
$D_6(x_1) = \range{1}{2}$, 
$D_6(x_2) = \range{0}{18}$, and
$D_6(x_3) = \range{0}{3}$
is not \ibounds{} consistent.
While the domain $D_7$ defined by
$D_7(x_1) = \range{1}{2}$, 
$D_7(x_2) = \range{1}{17}$, and
$D_7(x_3) = \range{2}{3}$
is \ibounds{} consistent,
supported by the integer solutions
$\{x_1 \mapsto 1, x_2 \mapsto 1, x_3 \mapsto 2\}$, and
$\{x_1 \mapsto 2, x_2 \mapsto 17, x_3 \mapsto 3\}$.
}
While this constraint is well-defined for integer values of the variables,
in order
to check \rbounds{} consistency 
we need to clarify what it means 
to be a \emph{real} solution of $x_1 = x_2 \mod x_3$.
\end{example}

\wh{Added another example}
\jeff{reworded last sentence}
\begin{example}
Consider the reified constraint $b \Leftrightarrow x_1 + x_2 \leq x_3$,
where $b = 1$ corresponds to $x_1 + x_2 \leq x_3$ being true, and $b = 0$
corresponds to it being false.
However, it is not clear what natural interpretation can be given to, say,
$b = 0.5$.
\end{example}

\section{Conditions for Equivalence}\label{sec:equiv}

Why has the confusion between the various
definitions of bounds consistency not been noted before? 
In fact, for many constraints, the definitions are \emph{equivalent}.

\ignore{ 
\begin{lemma}\label{lemma:ineq}
Let $c \equiv \sum_{i=1}^n a_i x_i \leq a_0$.
Then \rbounds{} and \ibounds{} consistency for $c$ is equivalent.
\end{lemma}
\noproofs{
\begin{proof}
The \rbounds{} and domain propagators for
these constraints are identical (see Schulte \& Stuckey~\cite{SchulteStuckey-PPDP2001})
hence the result holds. 
\end{proof}
}
} 


Following the work of Zhang \& Yap~\cite{ZhangYap00} 
we define $n$-ary monotonic constraints as a 
generalization of linear inequalities $\sum_{i=1}^n a_i x_i \leq a_0$.
Let $\theta \in_{\RR} D$ denote that $\theta(v) \in \RR$ and
$\inf_D v \leq \theta(v) \leq \sup_D v$ for all $v \in \vars(\theta)$.

\begin{defn}\label{defn:our-mon}
An $n$-ary constraint $c$ is \emph{monotonic with respect to variable
$x_i \in \vars(c)$} iff there exists a total ordering $\prec_i$ on
$D(x_i)$ such that if $\theta \in_{\RR} D$ is a real solution of $c$,
then so is any $\theta' \in_{\RR} D$ where $\theta'(x_j) = \theta(x_j)$ 
for $j \neq i$ and $\theta'(x_i) \preceq_i \theta(x_i)$.

An $n$-ary constraint $c$ is \emph{monotonic} iff $c$ 
is \emph{monotonic with respect to all variables in $\vars(c)$.}
\end{defn}

The above definition of
monotonic constraints is equivalent to but simpler than
that of Zhang \& Yap~\cite{ZhangYap00}; 
see Choi {\em et al.}~\cite{monotonicity} for justification and explanation.
\wh{Noted that we're only considering two possible total orders for
monotonicity.}
In the following, we assume that $\prec_i$ is $<$ or $>$ for all $i$, in
order to restrict ourselves to orders which are sensible with respect to
bounds consistency.

Examples of such monotonic constraints are:
\begin{itemize}
\item all linear inequalities
\item $x_1 \times x_2 \le x_3$
with non-negative domains, e.g.\ $D(x_i) = \range{1}{1000}$
for $1 \le i \le 3$.
\end{itemize}
For this class of constraints, \rbounds{}, \ibounds{} and \dbounds{}
consistency are equivalent to domain consistency. 

\begin{proposition}\label{lemma:monotone}
Let $c$ be an $n$-ary monotonic constraint.
Then \rbounds{}, \ibounds{}, \dbounds{} and domain 
consistency for $c$ are all equivalent.
\end{proposition}
\noproofs{
\begin{proof}
Since domain consistency $\Rightarrow$ 
\dbounds{} consistency $\Rightarrow$ 
\ibounds{} consistency $\Rightarrow$
\rbounds{} consistency, 
it is sufficient to show here that \rbounds{} consistency 
implies domain consistency for this constraint; 
thus we assume $D$ is \rbounds{} consistent and prove it is also domain
consistent.

\wh{Re-worked the proof}
Consider the greatest value $d_i \in D(x_i)$ for some $x_i \in \vars(c)$
(i.e.\ $d_i$ must be either $\inf_D x_i$ or $\sup_D x_i$ depending on the
ordering $\prec_i$ that makes $c$ monotonic w.r.t.\ $x_i$).
Since $D$ is \rbounds{} consistent, there exists $\theta \in_{\RR} D$ with
$\theta(x_i) = d_i$ such that $\theta$ is a real solution of $c$.

Let $\theta'(x_i) = \theta(x_i)$ and $\theta'(x_j) = d_j$ for all $j \neq i$
where $d_j$ is the smallest element of $D(x_j)$ with respect to $\prec_j$.
Since $c$ is monotonic, $\theta'$ is a solution of $c$, and since the
smallest elements of the domains are necessarily integers, $\theta'$ must be
an integer solution of $c$.

Now consider $\theta''$, where $\theta''(x_j) = \theta'(x_j)$ for all
$j \neq i$ and $\theta''(x_i) = d$ for some $d \in D(x_i)$.
Since $c$ is monotonic and $d \preceq_i \theta'(x_i)$ (since $\theta'(x_i)$
is the largest element of $D(x_i)$), $\theta''$ is also an integer solution
of $c$, with $\theta''(x_i) = d$.

Since the choice of $x_i$ and $d$ was arbitrary, we have shown that we can
construct an integer solution supporting any value in the domain of any
variable; thus $D$ is domain consistent.
\end{proof}
}

Although disequality constraints are not monotonic, they are
equivalent for all the forms of bounds consistency because they
prune so weakly. 

\begin{proposition}\label{lemma:disequal}
Let $c \equiv \sum_{i=1}^n a_i x_i \neq a_0$.
Then \rbounds{}, \ibounds{} and \dbounds{} consistency for $c$ are equivalent.
\end{proposition}
\noproofs{
\begin{proof}
(Sketch) Essentially if a value $d$ of a variable $x_i$ where $1 \le i \le n$ 
is supported by $c$, it is supported by an integer value in $D(x_i)$. 
Since either the domain of a variable is a singleton, 
in which case the bounds consistency must use the unique
integer, or the domain is not a singleton, and one
of the two integral endpoints will suffice. 
\end{proof}
}

All forms of bounds consistency are also equivalent
for binary functional constraints, such as 
$a_1 x_1 + a_2 x_2 = a_0$, $x_1 = a {x_2}^2 \wedge x_2 \geq 0$,
or $x_1 = 1 + x_2 + {x_2}^2 + {x_2}^3 \wedge x_2 \geq 0$.

\begin{proposition}\label{lemma:bij}
Let $c$ be a constraint with $\vars(c) = \{x_1, x_2\}$,
where $c \equiv x_1 = g(x_2)$ and
$g$ is a bijective and monotonic function.
Then \rbounds{}, \ibounds{} and \dbounds{} consistency for $c$ are equivalent.
\end{proposition}
\noproofs{
\begin{proof}
Assume $D$ is \rbounds{} consistent w.r.t.\ $c$.
Let us examine the endpoints of the ranges of each variable.
We will examine $d_1 = \inf_D x_1$, 
and assume $g$ is monotonically increasing. The other cases are similar.

Now $\{x_1 \mapsto d_1, x_2 \mapsto g(d_1)\}$ is the unique
real solution of $c$ supporting $d_1$.
Hence $\inf_D x_2 \leq g(d_1) \leq \sup_D x_2$.
Suppose that $g(d_1) > \inf_D x_2 = d_2$. 
Then $d_1 = g^{-1}(g(d_1)) > g^{-1}(d_2)$ since $g^{-1}$ is 
also monotonically
increasing.
Hence $d_2$ cannot be supported by any value in the domain of $x_1$.
Contradiction. Thus $d_2 = g(d_1) = \inf_D x_2$. 
Both $d_1 \in D(x_1)$ and $d_2 \in D(x_2)$ are integers 
since they are bounds.
Hence $\{x_1 \mapsto d_1, x_2 \mapsto g(d_1)\}$ is
an integer solution supporting $d_1$ for $x_1$.
Since the supporting solution only involves the end points 
in this case \ibounds{} and \dbounds{} are equivalent.
\end{proof}
}

\noproofs{
The proofs 
of Propositions~\ref{lemma:monotone},~\ref{lemma:disequal}, and~\ref{lemma:bij}
are related to but different from the proofs of
endpoint-relevance for these constraints 
in Schulte \& Stuckey~\cite{SchulteStuckey-PPDP2001}.

} 
\ignore{ 
\begin{proposition}\label{lemma:range-preserv}
Let $c \equiv \sum_{i=1}^n a_i x_i = a_0$, 
where $|a_i| = 1$ for $1 \leq i \leq n$.
Then \rbounds{} and \ibounds{} consistency for $c$ is equivalent.
\end{proposition}
\noproofs{
\begin{proof}
Lemma 2.23 of Schulte \& Stuckey~\cite{SchulteStuckey-PPDP2001} show that
the \rbounds{} propagator $\bnd(c)$ is such that
$\bnd(c)(D) \bequiv \dom(c)(D)$ for any range
domain $D$, which ensures \ibounds{} consistency.
\end{proof}
}
}

For linear equations with at most one non-unit coefficient, 
we can show that \rbounds{} and \ibounds{} consistency are equivalent.

\begin{proposition}\label{lemma:range-preserv2}
Let $c \equiv \sum_{i=1}^n a_i x_i = a_0$,
where $|a_i| = 1, 2 \leq i \leq n$, $a_0$ and $a_1$ integral.
Then \rbounds{} and \ibounds{} consistency for $c$ are equivalent.
\end{proposition}
\noproofs{
\begin{proof}
\Ibounds{} consistency implies \rbounds{} consistency, so it is sufficient
to show here that \rbounds{} consistency implies \ibounds{} consistency for
this constraint; thus we assume $D$ is \rbounds{} consistent and prove it is
also \ibounds{} consistent.

First consider a bound of $x_1$, say $d_1 = \inf_D x_1$ (the other bound
follows analogously).
Since $D$ is \rbounds{} consistent, there exist real
$d_i, \inf_D x_i \leq d_i \leq \sup_D x_i, 2 \leq i \leq n$
such that $\{x_i \mapsto d_i ~|~ 1 \leq i \leq n\}$ is a real solution of $c$.
Note that $\sum_{i=2}^n a_i d_i$ is integral, and since $|a_i| = 1$
($2 \leq i \leq n$), it is straightforward to construct integral $d_i'$ with
the same sum that respect the variables' bounds.\footnote{
    For example, compute the average fractional part of the non-integral
    $d_i$s, and then round that fraction of them up to the next integer; the
    rest are rounded down.
    This clearly respects the variable bounds, because they are integral and
    we never change a value beyond an integer.
} 
These $d_i'$ exhibit that the bound $d_1$ is \ibounds{} consistent.

Note that the fact that $x_1$ is \rbounds{} consistent also implies
\begin{equation}\label{eq:rbnd-impl}
\inf_D a_1 x_1 \geq a_0 - \sum_{i=2}^n \sup_D a_i x_i
\end{equation}

Now consider a bound of any of the other variables.
For simplicity we assume $a_2 = 1$ and consider the upper bound of $x_2$;
the other cases follow analogously.
Let $e_2 = \sup_D x_2$.
Since $D$ is \rbounds{} consistent, there exist real
$e_i, \inf_D x_i \leq e_i \leq \sup_D x_i, i \in \{1, 3 \ldots n\}$
such that $\{x_i \mapsto e_i ~|~ 1 \leq i \leq n\}$ is a real solution of $c$.
In particular, this means that
\[
\sup_D x_2 \leq a_0 - \inf_D a_1 x_1 - \sum_{i=3}^n \inf_D a_i x_i
\]
but~(\ref{eq:rbnd-impl}) implies
\[
\sup_D x_2 \geq a_0 - \inf_D a_1 x_1 - \sum_{i=3}^n \sup_D a_i x_i
\]
This means $\sup_D x_2 = a_0 - \inf_D a_1 x_1 - k$ where $k$ is an integer
and
$\sum_{i=3}^n \inf_D a_i x_i \leq k \leq \sum_{i=3}^n \sup_D a_i x_i$.
Clearly we can find integer values $f_i$ to assign to $x_i$
($3 \leq i \leq n$) to make their sum equal to $k$ while satisfying the
variable bounds, since the corresponding $a_i$ are $\pm 1$.
If we let $f_2 = \sup_D x_2$ and let $f_1$ be such that
$a_1 f_1 = \inf_D a_1 x_1$, then
$\{x_i \mapsto f_i, 1 \leq i \leq n\}$ is an integer solution of $c$ that
exhibits that $\sup_D x_2$ is \ibounds{} consistent.
\end{proof}
}


Even for linear equations with all unit coefficients,
\dbounds{} consistency is different from \ibounds{} and \rbounds{}
consistency.

\begin{example}[Counter example]
Consider the linear equation  $c \equiv x_1 + x_2 + x_3 = 5$ and the 
domain $D_8$ where 
$D_8(x_1) = \range{0}{3}$, $D_8(x_2) = D_8(x_3) = \{0,3,4,5\}$.
$D_8$ is \rbounds{} and \ibounds{} consistent as given by 
the following solutions of $c$:
$\{ x_1 \mapsto 0, x_2 \mapsto 0, x_3 \mapsto 5 \}$,
$\{ x_1 \mapsto 0, x_2 \mapsto 5, x_3 \mapsto 0 \}$, and
$\{ x_1 \mapsto 3, x_2 \mapsto 1, x_3 \mapsto 1 \}$.
$D_8$ is, however, not \dbounds{} consistent, 
since there is clearly no integer solution with $x_1 \mapsto 3$
(i.e.\ neither $1$ nor $2$ are in the domains of $x_2$ and $x_3$).
\end{example}

In summary for many of the constraints commonly used the notions
of bounds consistency are equivalent, but clearly not for all.

\section{Related Work}\label{sec:related}

In this paper we restricted ourselves to \emph{integer} constraint solving. 
Definitions of bounds consistency for \emph{real} constraints 
are also numerous, but their similarities and differences
have been noted and explained by  
e.g. Benhamou \bgroup \em et al.\egroup~\cite{newton}.
Indeed, we can always interpret integers as reals
and apply bounds consistency for real constraints plus 
appropriate rounding, e.g.\ CLP(BNR)~\cite{clpbnr}.
However, as we have pointed out in Section~\ref{sec:bc}, 
there exist integer constraints for which propagation is
less meaningful when interpreted as reals.

\jeff{reworded}
\ignore{
Lhomme~\cite{Lhomme93} defines \emph{arc B-consistency} which 
formalizes bounds propagation for both integer and real constraints.
He proposes an efficient propagation algorithm implementing
arc B-consistency with complexity analysis and experimental results. 
However, his study focuses on constraints defined 
by numeric relations (i.e.\ numeric CSPs).
}
Lhomme~\cite{Lhomme93} defines \emph{arc B-consistency} which 
formalizes bounds propagation techniques for numeric CSPs.
Unlike our definition of CSPs, constraints in numeric CSPs 
cannot be given extensionally and must be defined by numeric relations, 
which can be interpreted in either the real or the finite integer domain.
Numeric CSPs also restrict the domain of variables to be a single interval.

Walsh~\cite{toby-rc} introduces several new forms of bounds consistency
which extends the notion of $(i,j)$-consistency 
and relational consistency.
He gives a theoretical analysis comparing the propagation strength
of these new forms of bounds consistency.

Maher~\cite{prop-comp} introduces the notion of propagation completeness
together with a general framework to unify a wide range of 
consistency. 
These include hull consistency of real constraints
and \ibounds{} consistency of integer constraints.
Propagation completeness aims to capture the timeliness 
property of propagation.

The application of bounds consistency is not limited to integer
and real constraints. 
Bounds consistency has been formalized for solving 
set constraints~\cite{conjunto}, and more recently, 
multiset constraints~\cite{multiset-cp03}.

\section{Conclusion}\label{sec:concl}

The contributions of this paper are two-fold.
First, we point out that the three commonly used definitions 
of bounds consistency are incompatible.
Second, we clarify their differences and show that for 
several types of constraints, \rbounds{}, \ibounds{} 
and \dbounds{} consistency are equivalent.
This explains partly why the discrepancies among the definitions
were not noticed earlier.
Our precise definitions can serve as the basis for 
verifying all implementations of bounds propagators.

\ignore{
An important direction for future work is 
to build efficient \rbounds{} and/or \ibounds{} propagators 
for new classes of constraints.
\Dbounds{} consistency resembles much of domain consistency,
yet only makes guarantees about the endpoints of the domain.
It would be interesting to determine whether there are some CSPs
where \dbounds{} propagators would be advantageous.
}

\bibliographystyle{abbrv}
\bibliography{paper}

\end{document}